# Robustness Evaluation of Two CCG, a PCFG and a Link Grammar Parsers

**Tuomo Kakkonen**

Department of Computer Science and Statistics, University of Joensuu
P.O. Box 111, FI-80101 Joensuu, Finland
tuomo.kakkonen@cs.joensuu.fi

**Abstract**

*Robustness* in a parser refers to an ability to deal with exceptional phenomena. A parser is *robust* if it deals with phenomena outside its normal range of inputs. This paper reports on a series of robustness evaluations of state-of-the-art parsers in which we concentrated on one aspect of robustness: its ability to parse sentences containing misspelled words. We propose two measures for robustness evaluation based on a comparison of a parser's output for grammatical input sentences and their noisy counterparts. In this paper, we use these measures to compare the overall robustness of the four evaluated parsers, and we present an analysis of the decline in parser performance with increasing error levels. Our results indicate that performance typically declines tens of percentage units when parsers are presented with texts containing misspellings. When it was tested on our purpose-built test set of 443 sentences, the best parser in the experiment (C&C parser) was able to return exactly the same parse tree for the grammatical and ungrammatical sentences for 60.8%, 34.0% and 14.9% of the sentences with one, two or three misspelled words respectively.

## Introduction

This paper investigates parser evaluation from the point of view of *robustness*. It is a part of a set of evaluations in which selected parsers are evaluated using five different criteria: *preciseness*[1], *coverage*, robustness, *efficiency* and *subtlety*. Parsers are often presented with texts that contain errors. Thus, for example, a parser, while processing user inputs, may encounter misspelled words, incorrect cases, missing or extra words, or dialect variations. Transcriptions of *spoken* language texts are especially likely to contain such errors and complications. A parser's ability to produce an error-free or only a slightly altered output from input sentences containing errors is referred as robustness. A robust parser is able to recover from various types of exceptional inputs and provide as complete and correct analysis of the input sentence as it is capable of doing under the circumstances. A total failure on the part of the parser to produce an output is only accepted when the input text is extremely distorted or disfigured.

Even though a high degree of robustness is regarded as a desirable characteristic in a parser,[2] most parser evaluation protocols and campaigns tend to ignore robustness as a criterion for evaluation. Most evaluations focus on measuring the preciseness of the structures assigned by parsers. Evaluating the preciseness of a parser's output consists of making judgments about the grammaticality or the correctness of structural descriptions assigned by the system. This kind of parser evaluation compares a system output to the correct human-constructed parses obtained from a treebank.

We have carried out a series of robustness evaluations for English parsers that are, as far as we know, the only evaluations of their kind to be reported in literature. In our evaluation of the robustness of four state-of-the-art parsers, we sought answers to the following questions:

- What is the overall robustness of these state-of-the-art parsers?
- What effect does an increasing error level have on the parsing results?
- Is there a connection between the preciseness and robustness of these parsers? In other words, is robustness achieved at the price of overgeneration that creates disambiguation problems?

## Background

### Robustness, overgeneration, efficiency and dealing with ill-formed input

There is an important connection between robustness and *overgeneration*. If robustness is achieved by adding new rules to the grammar and/or relaxing the constraints, it is highly likely that a parser thus defined will suffer from overgeneration and produce huge numbers of candidate parses for every sentence (even ungrammatical ones). This, in turn, would diminish the preciseness of the system. Robustness may also affect the efficiency of a parser because robustness mechanisms often generate more processing and so overload a system's computational capacity.

Probabilistic approaches to parsing are inherently robust because they consider all possible analyses of a sentence and always propose a parse for any given input. Robustness can be added to a symbolic, rule-based parser in several ways (Nivre, 2006; Menzel, 1995). The first is to relax the constraints of the grammar in such a way that a sentence outside the language generated by the grammar can be assigned a complete analysis. The second is to get a parser to try to recover as much structure as possible from the well-formed fragments of an analysis when a complete analysis cannot be performed. The third is to identify a number of common mistakes and integrate them into the grammar in anticipation of such errors occurring in texts. If one used the third method, one would have to limit oneself to few predictable high-frequency kinds of errors such as common spelling mistakes and mistakes in word order.

---

[1] Instead of the frequently used term "accuracy", we use the term "preciseness" to refer to the correctness of analyses assigned by a parser. We avoid using the term "accuracy" because of its technical use in evaluation context. *Test accuracy* typically refers to the fraction of instances correctly classified. It is therefore logical to use the term "accuracy" to refer to the percentage of constituents/dependencies or sentences correctly parsed.

[2] Since some parsers are designed to check grammar, their returning of a "failure to parse" in response to an ungrammatical sentence is (for them) a "correct" result. The evaluation approach discussed in this paper is not applicable to such systems. In such cases, one could use the proportion of ill-formed sentences that the parser accepts as a measure of robustness.

## Previous work on evaluation of robustness against ill-formed input

Relatively little work has been done on methods of empirically evaluating the robustness of parsers. Foster (2004), for example, describes a resource for evaluating a corpus of ungrammatical English sentences. The error types in her corpus include incorrect word forms, extraneous words, omitted words, and composite errors. For each sentence, the corpus offers parallel correct and ungrammatical versions that are identical in meaning. A comparison of the parsers' output for well-formed and ill-formed inputs is used for evaluating the robustness of a parser when it is confronted with ill-formed input sentences. Foster's resource is, however, not relevant to this present research because we focus solely on robustness in relation to misspelled words.

The research most similar to our own is that reported by Bigert et al. (2005) who manufactured ill-formed sentences for the purpose of evaluating robustness by using an automatic tool (that simulates naturally occurring typing errors) to introduce spelling errors into input sentences. The automatic introduction of errors thus produced enabled the researchers to undertake a controlled testing of *degradation*, the effect of an increased error rate on a parser's output. Evaluation occurs in the following way: Firstly, the parser to be evaluated is given an error-free text to parse. Secondly, the parser is given ill-formed input texts to parse. Finally, the results obtained from first and second stages are compared. The degradation of a parser's output is then measured by comparing the parser's preciseness on error-free texts to its preciseness on ill-formed inputs. The procedure is repeated for several levels of distortion (examples being 1%, 2% and 5% of the input words). But Bigert et al. applied their method only to part-of-speech (POS) taggers and parsers for Swedish.

## Experiments

In order to accomplish our purpose of evaluating the degree to which a parser can handle noisy input (and spelling errors in particular), we set up the following experiment. Our first need was a set of test sentences, both correct and erroneous. But because no annotated evaluation resources with erroneous sentences exists for testing parsers, we had to approach the problem in a different way. We in fact adopted a method similar to that of Bigert et al. (2005) and utilized a comparison between the analyses that the parser produced for the correct and for the noisy versions of sentences. We therefore did not need an annotated "gold standard" for each parser. Our second need was to find an evaluation metric and measures wherewith to compare the performances of the systems. For this purpose we devised a metric and two measures. We then considered the preciseness figures reported in the literature for the four parsers and compared them to our findings.

## Parsers

We chose four state-of-the-art parsers for this experiment. They use three different grammar formalisms: *Combinatory Categorial Grammar* (CCG) (Steedman, 2000), *Probabilistic Context-free Grammar* (PCFG) (Booth & Thompson, 1973) and *Link Grammar* (LG) (Sleator and Temperley, 1991). We ran all the experiments on the default settings of the parsers and (depending on the parser) either on Linux or on SunOS or on a Windows XP environment.

### C&C Parser

The *C&C Parser* (v. 0.96, 23 November 2006) is based on a CCG (Clark and Curran, 2004). It applies log-liner probabilistic tagging and parsing models. This parser uses a lexical category set containing 425 different categories and produces an output that contains CCG predicate-argument dependencies and *grammatical relations* (GR) output, as defined by Carroll et al. (2003). We used the GR output for evaluation. The format consists of 48 relation types that indicate the type of dependency that exists between the words. The preciseness of the system on *Penn Treebank* (PTB) (Marcus et al. 1993) data is reported at 86.6% and 92.1% for labeled precision and recall respectively.

### Link Grammar Parser

The *Link Grammar Parser* that arose out of the work of Sleator and Temperley (1991), is based on a dependency grammar-style formalism.[3] We used the version 4.1b (January 2005) of the system in the experiment. The search algorithm uses a dynamic programming approach. While this parser can produce both phrase structure and LG analysis, we used the latter for evaluation purposes. These so-called *linkages* consist of undirected links between words. The syntactic tagset consists of 107 link types. The system outputs several analyses for the sentences to which more than one plausible linkage can be found. This occurred with most of our test sentences. But in the interests of ensuring an entirely fair comparison among the parsers, we considered only the first, highest-ranking linkage in the evaluation. Molla and Hutchinson (2003) report 54.6% precision and 43.7% recall on recovering GR links (only four link types (subj, obj, xcomb, mod) were included) on a test set consisting of 500 sentences from SUSANNE corpus (Sampson 1995).

### Stanford Parser

In contrast to most other parsers based on PCFGs, the *Stanford Parser* (v. 1.5.1, 30 May 2006) – referred in the remainder of this text as SP – is based on an unlexicalized model (Klein and Manning, 2003). This parser uses a a *Cocke-Younger-Kasami* (CYK) (Kasami, 1965, Younger, 1967) search algorithm and can output both dependency and phrase structure analyses (de Marneffe et al., 2006). We ran the experiment on the English PCFG grammar and then carried out the evaluation on the dependency-style output (consisting of 48 GRs) in the same way that we carried out the evaluation on the C&C parser. We considered only the highest-ranking parse for each sentence. Klein and Manning (2003) reported labeled precision and recall figures of 86.9% and 85.7% respectively for this parser.

---

[3] The main differences are as follows: (1) Links in LG are undirected. (2) Links may form cycles. (3) There is no notion of the root word. In addition, Link Grammar is fully projective and context-free. This is in contrast to what one finds in many dependency grammars.

**StatCCG**

StatCCG (Preliminary public release, 14 January 2004) is a statistical parser for CCG that was developed by Julia Hockenmaier (2003). StatCCG was the only parser in this experiment that was not able to perform POS tagging. We therefore used the MXPOS tagger (Ratnaparki, 1996) to preprocess the texts before inputting them to StatCCG. In contrast to C&C, this parser is based on a generative probabilistic model. Its lexical category set contains approximately 1,200 types, and there are four atomic types in the syntactic description. The parser produces two types of output: the CCG derivations, and the word-word dependencies in the predicate-argument structure. We used the latter format for our evaluation. The reported precision and recall of the parser on PTB data is 90.5% / 91.1% over unlabelled and 83.7% / 84.2% over labeled dependencies respectively.

**Test materials**

An annotated evaluation resource would be needed to measure the robustness of a parser against human judgments. It would, however, be a daunting task to annotate noisy texts and their corresponding correct counterparts for all four of the parsers. We therefore made the simplifying assumption, like (Bigert et al. 2005), that a parser is robust if it is able to produce a similar analysis for a correct sentence and a noisy version of the same sentence. Our assumption is that if a parser is able to do this, it will be able to perform in a robust way when it is confronted by noisy inputs. By making this assumption, we were therefore able to perform evaluations by using unannotated texts.

It is clear that as the level of noise in the inputs increases, the performance of a system degrades correspondingly. The extent to which this occurs can be measured by increasing the number of mistakes in the input sentences and observing the effect that this has on its performance. In order to investigate the effect of increasing the amount of distortion in the input, and to answer our second research question, we constructed a test corpus which contained sentences with error-free sentences and their noisy counterparts with one or more spelling errors. Our test set had three error levels: each of the noisy sentences contained between one to three misspelled words.

We started the test set construction by selecting 19 sentences from a public domain web page. We then altered one, two or three words per test sentence and this gave us a total of 443 test sentences – 255 with one error and 94 with two and three errors respectively. The length of each of these sentences was between 5 and 36 words and the average length was 16.32 words per sentence. We then introduced misspellings manually into the sentences by deleting, adding and swapping characters, permitting only one edit operation per word.

We based character additions on the keyboard proximity of letters in order to simulate errors in naturally-occurring texts. Since our purpose was not to evaluate robustness on structurally distorted sentences, we only permitted alterations that did not create an acceptable (valid) word.

**Test settings**

We carried out evaluation of the parsers in the following way: First, we parsed the correct sentences (CS). Secondly, we parsed the three sub-corpora of noisy sentences, each representing an error-level from one to three ($NS_1$, $NS_2$, $NS_3$). Thirdly, we implemented an evaluation tool that automatically compared the analyses produced for each sentence in CS and its corresponding sentence in $NS_1$, $NS_2$ and $NS_3$ respectively. Finally, we metered the performance by using two distinct evaluation measures. The first place, we calculated the number of sentences for which a parser produced exactly the same structure for both the correct and noisy input sentence. We refer to this measure as an *unlabeled robustness score*. Our second measure, *labeled robustness score*, is stricter: it accepts an analysis only if the two structures are the same and if, in addition, the labels on syntactic categories (GRs, dependencies) match. For example, the introduction of a single misspelling into a sentence often results the type of the dependency link associated with the misspelled word to be altered.

**Results**

Table 1 (below) summarizes the results of our experiments and reports on overall robustness scores as well as separate scores for each error level. Table 1 shows that C&C was the best overall performer in this experiment. The overall performance of StatCCG and SP are similar on the unlabeled evaluation in which the LG parser performs considerably worse than the other three parsers. The results also indicate (as we expected they would) that the performance of the parsers degrades as the level of distortion in the input sentences increases.

| Parser | Unlabeled similarity | | | | Labeled similarity | | | |
|---|---|---|---|---|---|---|---|---|
| | *Overall* | *1* | *2* | *3* | *Overall* | *1* | *2* | *3* |
| C&C | 63.88 | 72.94[*] | 62.77 | 40.43 | 45.37 | 60.78[*] | 34.04 | 14.89 |
| StatCCG | 57.11 | 72.55 | 41.49 | 30.85 | 44.02 | 58.82 | 27.66 | 20.21 |
| SP | 55.30 | 70.98 | 42.55 | 25.53 | 19.19 | 29.41 | 9.57 | 1.06 |
| LG | 29.80 | 40.39 | 22.34 | 8.51 | 17.61 | 21.96 | 20.21 | 3.19 |

Table 1. The results of the experiment. For both the metrics, we give the overall robustness score as a percentage of the accepted parses. We also give separate scores for each error level. [*]C&C failed to parse 23 correct sentences in this sub-corpus. Because we consider the failure of a parser to cover some sentences to be a serious robustness flaw, we included these sentences in the calculations. This brought down the scores from 80.17 to 72.94 and 66.81 to 60.78 for structural and structural and label similarities respectively.

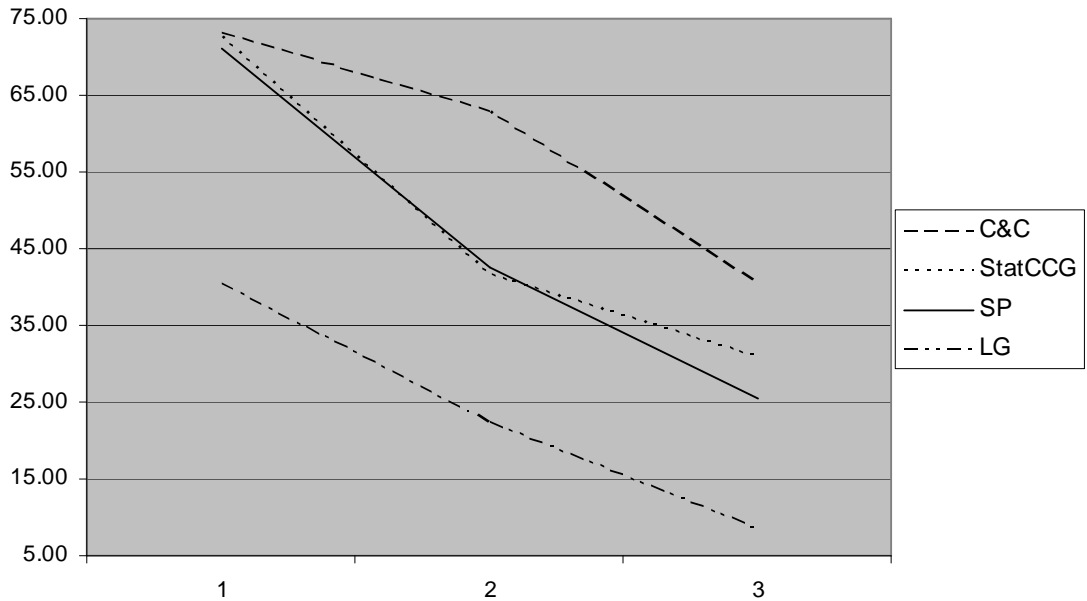

Figure 1. Unlabelled robustness scores on the three error levels. The y-axis indicates the robustness score on the three error levels given on the x-axis.

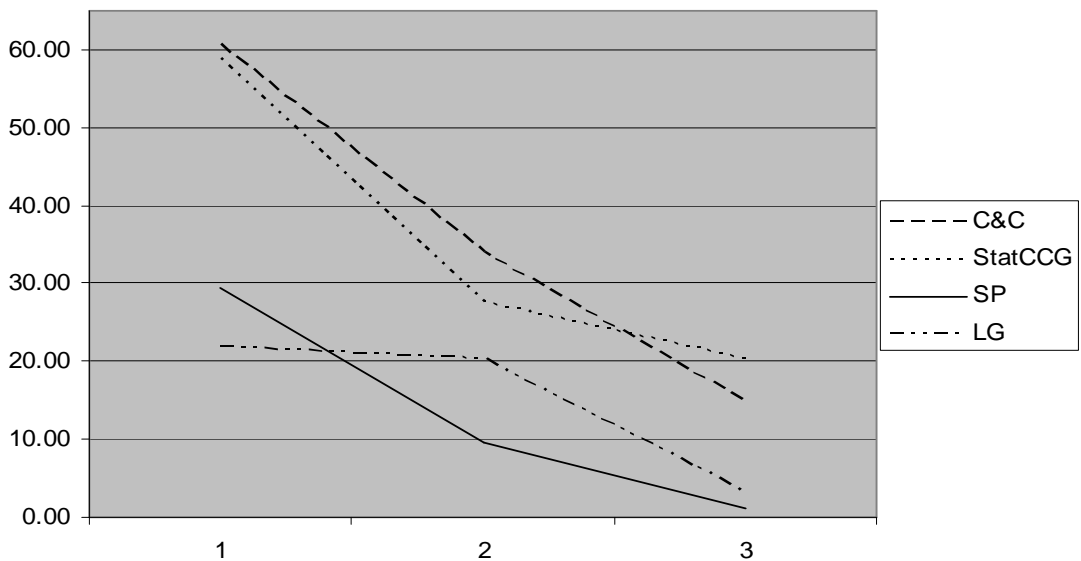

Figure 2. Labelled robustness scores.

Figures 1 and 2 (above) illustrate the performance of the parser on each error level for labeled and unlabeled respectively. The results of the labeled evaluation indicate that while structural similarity can be preserved for close to 73% of the sentences on error level 1 by the two best parsers (StatCCG and C&C), the performance drops to about 60% when the labels are also required to match. Error level 3 on unlabeled and labeled evaluation was the only category in which StatCCG outperformed C&C in this experiment. The large tagset of 107 tags makes it difficult for LG to obtain a good performance on labeled evaluation. It was rather surprising, however, to observe the drastic drop in SP's robustness scores. This indicates a flaw in the robustness mechanisms of the parser, a flaw that might be attributable to the poor ranking of candidate parse trees for noisy sentences or problems with the POS tagging model of unknown words.

We made an interesting observation by comparing the overall robustness scores to the preciseness of the systems measured by the F-score[4] over precision and recall figures reported in the literature and given above. The F-scores are 89.3 for C&C, 86.3 for SP, 83.9 for StatCCG and 48.5 for LG. This comparison suggests a correlation between the preciseness and robustness figures of the four parsers. In addition to being accurate, C&C is the most consistent parser when faced with distorted input: the unlabelled robustness score dropped 44.6% from error level 1 to error level 3. LG has the lowest reported preciseness and its performance also degrades drastically, especially on unlabelled evaluation (a 78.9% drop from error level 1 to error level 3). StatCCG and SP are once again in between these two extremes, scoring 57.5% and 64.0% lower on error level 3 respectively. On labeled evaluation, the order is as follows: StatCCG (65.5%), C&C (75.5%), LG (85.5%), and SP (96.4%). These figures support our earlier observation that StatCCG is the best performer on higher error levels and that SP's robustness mechanism contains flaws.

---

[4] We use the F-score definition $\frac{2*Precision*Recall}{Precision+Recall}$

## Conclusion

This paper describes our experiments in parsing texts with misspelled words. Our aim in conducting his experiment was to evaluate the robustness of four parsers based on three different grammar formalisms. C&C was the only parser in the experiment that had coverage less than 100% on our test set. In spite of this, it proved itself to be by far the best-performing parser in this research. In comparison to StatCCG, C&C performed better – especially on the unlabeled measure. SP did relatively well on recovering the structure of noisy input sentences, but performed worse than all the other parsers on labeled evaluation on error levels 2 and 3. LG's performance left a lot to be desired across all of the evaluation categories. LG also returned up to thousands of linkages per input sentence. All these observations, together with the low preciseness figures reported in literature, indicate serious problems in the disambiguation model of the parser.

Several interesting directions for future work suggest themselves in this field of parser evaluation. It would be interesting, for example, to extend this work with more parsing systems. In addition to collecting data for more comprehensive system comparisons, such experimentation would allow for deciding whether or not there exists a general correlation between the preciseness and the robustness of parsers. In addition to evaluating more parsers, this kind of research could be extended to include kinds of noise other than misspellings. A future researcher might, for example, use the corpus of Foster (2004) as a source for syntactically distorted sentences.